# A Summary of
# A New Normative Theory of Probabilistic Logic

*Romas Aleliunas*

*Simon Fraser University, Burnaby, British Columbia, Canada  V5A 1S6*

ABSTRACT       *By probabilistic logic I mean a normative theory of belief that explains how a body of evidence affects one's degree of belief in a possible hypothesis. A new axiomatization of such a theory is presented which avoids a finite additivity axiom, yet which retains many useful inference rules. Many of the examples of this theory—its models—do not use numerical probabilities.*

*Put another way, this article gives sharper answers to the two questions—*
*1. What kinds of sets can used as the range of a probability function?*
*2. Under what conditions is the range set of a probability function isomorphic to the set of real numbers in the interval [0,1] with the usual arithmetical operations?*

Any mechanical device (let's just call it a box) that makes decisions or performs actions in the real world can be analysed *as if* it were acting on the basis of—

    (1) a body of statistical knowledge, and

    (2) a system of utilities or preferences,

whether or not anyone can find, upon opening the box to study its mechanism, obvious embodiments of (1) or (2). Describing a box's decision-making behaviour in this way helps us see more quickly which real world situations the box can handle successfully, and which ones it cannot—mathematical statistics, and the method of performing experimental trials (on which statistics is based), are, after all, the only tools we have for making such judgements. (The difficulties with non-monotonic and default logic arise because we are in no position with these theories to make such judgements. Who knows when it is safe to say "Birds typically fly" unless this statement has a statistical interpretation?—something which is usually denied by the inventors of these theories.)

According to KEYNES, probability in its broadest sense, is the study of methods of encoding a body of statistical knowledge—item (1) above—as a set of likelihood relationships between evidence statements and hypotheses, or, more generally, between pairs of propositions. Typically this collection of statistical statements contains forumlae about conditional probabilities such as "p(A|B)=0.5".

KEYNES suggested that limiting this collection to inequality statements of the form $p(A|B) \geq p(C|D)$ is perhaps the weakest and least objectionable way to code statistical knowledge. This method does not require and does not introduce any specific numerical probability assignments, except in trivial cases such as $p(A|B) \geq p(A|A)$ which is merely a roundabout way of saying $p(A|B) = 1$. This idea of using only partial ordering to encode statistical knowledge was taken up again by B. O. KOOPMAN and studied in more depth in his paper of 1940. My goal here is to take pursue this idea once again from a more algebraic point of view.

The algebraic approach I take assumes that to each possible conditional probability, say p(A|B), there is assigned the formal probabilty (p,A,B), where (p,A,B) is just an uninterpreted formal mark. Thus these formal marks do not necessarily denote real numbers, for instance.

This maneuver allows us to replace inequalities between symbols of the form p(A|B) with inequalities between symbols of the form (p,A,B) which we will treat as atomic symbols. By atomic I mean that we cannot look into the internal structure of (p,A,B) to see that it is about the propositions A and B and the probability function p. Let us call the set of all possible formal marks P*. So far we have merely replaced one system for encoding statistical knowledge by a mathematically equivalent one. Our statistical knowledge is to be encoded now by the choice of partial ordering for P*.

The possible equivalence classes of P* are important. At one extreme we may have as many equivalence classes as there are elements in P*, and at the other extreme we may only have three equivalence classes. The first possibility corresponds to KEYNES's suggestion— what FINE [1973] calls a "weak comparative system of probability"—while the second one sorts likelihoods into three categories: "impossible", "certain", and "maybe". Yet another possibilty is to have equivalence classes correspond to real numbers and to deduce the inequalities between conditional probability symbols from the corresponding inequalities between real numbers.

This algebraic maneuver, therefore, gives us a uniform language for studying any sort of set of probabilities **P** that can ever be conceived. Suppose that some alien statistician chooses to

9

assign probabilities by drawing probability values from some partially ordered set **P**. Then the set **P** must be isomorphic to one of the possible systems of equivalence classes of P*.

Obviously the interesting question is what sorts of sets **P** are sensible choices for the range of a probabilty function such as p(·|·)? Further, under what conditions, assuming they meet some weak notion of rationality, must the range of a probability function be isomorphic to the real numbers in [0,1] under the usual arithmetical operations? (This second question was studied by Richard T. COX in 1946. He attempted to show that the answer is "always".) My aim here is to provide more precise answers to these two questions. (My answer to the second question turns out to be "not always".)

Aside from streamlining the mathematical study of abstract probability, the algebraic approach promises to be both more natural from a psychological point of view, and to be more attractive computationally. Both of these points are taken up more fully in another article I am preparing.

## What is Probabilistic Logic ?

Probabilistic logic is any scheme for relating a body of evidence to a potential conclusion (a hypothesis) in a rational way. We assign *degrees of belief* (which we also call *probabilities* ) to the possible relationships between hypotheses and pieces of evidence. These relationships are called *conditionals*. We will use the expression "f(P|Q)" to stand for "the conditional probability of P given the evidence Q as given by the probability assignment f." We have chosen to encode the potential hypothesis, P, and the collected evidence, Q, as finite sentences of some language **L**. We take *probabilistic logic* to be synonymous with *theory of rational belief* , *probability* to be identical to *degree of belief*, and *probability assignment* to be the same as *belief function* . Normative theories of rational belief have had a long history (KEYNES [1921], CARNAP [1950], FINE [1973]).

Let the set of probabilities be **P**. We begin by regarding this as a set of *uninterpreted formal marks*. A theory of probabilistic logic is chiefly concerned with identifying the characteristics of a family **F** of functions from **L**x**L** to **P**. **F** is the *set of permissable belief functions* (also called the *set of possible probability assignments* ) from which a rational agent is permitted to choose. Our convention is that for any f in **F**, f(P|Q) represents the probability assigned by f to the hypothesis P given the evidence Q. P and Q can be any statements in **L**—no distinction is made between hypothesis statements and evidence statements in probabilistic logic.

The set **F** of rational probability assignments clearly cannot be the set of all functions from **L**x**L** to **P**. Both **F**, and each individual element in it, must be subject to some constraints. Of course different theories of rational belief may vary in their choices for these constraints, but I believe they all share some in common. One such common constraint is: if "I believe P is certain when I see Q" then it cannot also be the case that "I believe ~P is possible when I see Q." Moreover, the correctness of these constraints—and therefore of the rules of inference they engender—cannot depend on the particular application that they are put to. Decision "theory" does not dictate the standards of correctness for inferences in probabilistic logic (TUKEY [1960]).

Does the set of probabilities **P** have any internal structure? We must, at the very least, be able to assert that some conditionals are more believable than others, or else there is no practical use for this theory. This implies that we have, at the very least, a partial ordering among conditionals. This algebraic approach does not lose us any generality because, as indicated in the introduction, we may introduce a new probability symbol into **P** for each pair of sentences drawn from the language **L** and assign whatever (partial) set of orderings between these formal symbols that we wish. In this way we can reproduce any rational partial ordering whatsoever between the conditionals as an ordering of the elements of **P**. A theory which adopts this approach is called a *weak comparative theory of probability* in FINE [1973].

A theory of rational belief is not a substitute for a decision-making procedure. Probabilistic logic does not prescribe "rules of detachment," "rules of commitment," or other decision-making rules. The only judgements a theory of probabilistic logic tells you how to make are conditional judgements—how to assign a value to a conditional probability. Conditional judgements of probability are the only judgements that can be infallible. Moreover, conditional judgements are the natural inputs to a decision-making apparatus.

## New Axioms for Probabilistic Logic

The probabilities in **P** are treated as uninterpreted formal marks in the axiomatization given below. We will exhibit ssome possible interpretations later.

These axioms describe the constraints that are appropriate for any rational theory of belief. The axioms fall into three groups: (1) axioms about the domain and range of each probability assignment f in **F**, (2) axioms stating consistency constraints that each individual f in **F** must obey, and (3) axioms that say something about **F** itself. Finite additivity does not appear as an axiom since it essentially forces probabilities to be numbers.

*AXIOMS for PROPOSITIONAL PROBABILISTIC LOGIC*

*Axioms about the domain and range of each f in **F**.*
1 . The set of probabilities, **P**, is a partially ordered set. The ordering relation is " $\leq$ ."

2. The set of sentences, **L**, is a free boolean algebra with operations &, v, and ~, and it is equipped with the usual equivalence relation " ≈ ." The generators of the algebra are a countable set of *primitive propositions*. Every sentence in **L** is either a primitive proposition or a finite combination of them. (See BURRIS *et al.* [1981] for more details about boolean algebra.)

3. If P ≈ X and Q ≈ Y, then f(P|Q)=f(X|Y).

*Axioms that hold for all  f in* **F**, *and for any* P, Q, R *in* **L**.

4. If Q is absurd (*i.e.* Q ≈ R&~R), then f(P|Q) = f(P|P).
5. f(P&Q|Q) = f(P|Q) ≤ f(Q|Q).
6. For any other g in **F**, f(P|P) = g(P|P).
7. There is a monotone non-increasing total function, i, from **P** into **P** such that f(~P|R) = i( f(P|R) ), provided R is not absurd.
8. There is an order-preserving total function, h, from **P**x**P** into **P** such that f(P&Q|R) = h(f(P|Q&R), f(Q|R)). Moreover, if f(P&Q|R) = 0*, then f(P|Q&R) = 0* or f(Q|R) = 0*, where we define 0* = f(~R|R) as a function of f and R.
9. If f(P|R) ≤ f(P|~R) then f(P|R) ≤ f(P|Rv~R) ≤ f(P|~R).

*Axioms about the richness of the set* **F**.

Let **1** = Pv~P. For any **distinct primitive** propositions A, B, and C in **L**, and for any arbitrary probabilities a, b, and c in **P**, there is a probability assignment f in **F** (not necessarily the same one in each case) for which—

10. $\quad$ f(A|**1**) = a,  f(B|A) = b, and f(C|A&B) = c.
11. $\quad$ f(A|B) = f(A|~B) = a and f(B|A) = f(B|~A) = b.
12. $\quad$ f(A|**1**) = a and f(A&B|**1**) = b, whenever b ≤ a.

Explanatory comments about these axioms can be found in the complete article. Axiom 7 has a precondition that did not appear in the earlier article.

## Possible Interpretations for the Set P

One can show the existence of well-defined probabilities called 0 and 1 in any set **P** that satisfies these axioms. To give a mathematical model for a **P** which satisfies these axioms we must, therefore, give a consistent interpretation to each component of the structure (**P**, ≤, h, i, 0, 1).

*(Model 1)   0,1-Probabilistic Logic*
$P(2) = \{0, 1\}$ and $0\cdot 0 = 0\cdot 1 = 0$, $1\cdot 1 = 1$, $i(0) = 1$, $i(1) = 0$, and $0 \leq 1$.

*(Model 2)   Simple Real Probabilistic Logic*
$P(R) = [0,1]$, the closed interval of real numbers from 0 to 1. Let $p\cdot q$ be the ordinary numerical product, and let $i(p) = 1 - p$. Use the usual ordering relation.

For more examples suppose $(P, \leq)$ is a totally ordered set with n elements. Then the table below displays the number, $N(n)$, of non-isomorphic models consistent with this constraint—

| n | 2 | 3 | 4 | 5 | 6 | 7 |
|---|---|---|---|---|---|---|
| N(n) | 1 | 1 | 2 | 3 | 7 | 16 |

Yet another model of these axioms is the following one mimicing English probabilities. My aim here is not to claim that this is the correct model for English probabilities, but instead to show that a good approximation to English probabilities will not be ruled out by these axioms.

*(Model 3)   Simplified English Probabilistic Logic*
Let $P(E)$ be the algebra of probabilities generated by the two formal symbols {LIKELY, UNLIKELY}, subject to the following additional constraints:
(C1)    UNLIKELY = i(LIKELY)
(C2)    0 < UNLIKELY < LIKELY < 1.
The elements of $P(E)$ are strings of symbols generated by—
(G1)    concatenating previously constructed strings in $P(E)$,
(G2)    forming i(s) where s is a string in $P(E)$,
(G3)    introducing the formal symbol s(p,q) for any pair of elements p and q in $P(E)$ whenever $p \leq q$ and there does not yet exist a solution, r, in $P(E)$ for the equation $p = q\cdot r$.

The only ordering relations are those that can be inferred from (C1), (C2), and the properties of the functions h and i. This set is not totally ordered. The symbols introduced by (G3) do not, as far as I know, have English names. But not being able to name them in English does not mean they do not exist in principle.

## When Must P Be Like The Real Numbers?
Under what conditions does the familiar numerical theory of probability with the familiar arithmetical operations become the only possible theory of rational belief? Richard COX [1946, 1961] first investigated this question under the assumption that probabilities were real numbers and concluded that the algebraic operations used must then be the same as the ordinary ones. ACZEL [1966] reports stronger results under the assumption of finite additivity for probabilities. The following theorem is in the same vein, but it drops these two assumptions and even some of the axioms. I call it the Tar-Pit Theorem because it shows how easy it is to get stuck in stuff that has been lying around for a long time.

*Definition*

**P** is *archimedean ordered* if, for any $p \neq 1$ and any $q \neq 0$ in **P**, the repeated product $p \cdot p \cdot \ldots \cdot p = p^n$ becomes smaller than q for some positive integer n.

*The Tar-Pit Theorem*

Under the conditions established by axioms 1 through to 12, the following two statements are logically equivalent:
(1) The set of probabilities **P** is totally ordered and also archimedean ordered.
(2) The algebraic structure $(\mathbf{P}, \geq, h, i, 0, 1)$ is isomorphic to a subalgebra of the system $(\mathbf{P}(\mathbf{R}), \geq, \cdot, i, 0, 1)$ given as Model 2, namely Simple Real Probabilistic Logic.
Dropping both axiom 11 and the assumption that products have no non-trivial zeroes (a part of axiom 8) does not affect the theorem.

A sketch proof of this is given in the CSCI article.